\documentclass[10pt,twocolumn,letterpaper]{article}

\usepackage[pagenumbers]{cvpr} 

\usepackage{graphicx}
\usepackage{amsmath}
\usepackage{amssymb}
\usepackage{booktabs}

\usepackage{color, colortbl}
\definecolor{LightCyan}{rgb}{0.88,1,1}
\usepackage{algorithm}
\usepackage{algorithmic}
\usepackage{multirow}
\usepackage{verbatim}
\usepackage{color}
\usepackage{gensymb}
\usepackage{amsmath}
\usepackage{subfiles}
\usepackage{bbm}
\usepackage[accsupp]{axessibility}  
\usepackage{amssymb}  

\usepackage[switch]{lineno}

\usepackage[pagebackref,breaklinks,colorlinks]{hyperref}

\usepackage[capitalize]{cleveref}
\crefname{section}{Sec.}{Secs.}
\Crefname{section}{Section}{Sections}
\Crefname{table}{Table}{Tables}
\crefname{table}{Tab.}{Tabs.}

\def\cvprPaperID{8646} 
\def\confName{CVPR}
\def\confYear{2022}

\begin{document}

\title{Boosting 3D Object Detection by Simulating Multimodality on Point Clouds}

\author{Wu Zheng$^{1}$ \quad Mingxuan Hong$^{1,2}$ \quad Li Jiang$^{3}$ \quad Chi-Wing Fu$^{1,2}$ \\
$^{1}$Department of Computer Science and Engineering, CUHK \\ $^{2}$Shun Hing Institute of Advanced Engineering, CUHK \hspace*{1cm} $^{3}$Max Planck Institute\\
{\tt\small \{wuzheng, cwfu\}@cse.cuhk.edu.hk\quad lijiang@mpi-inf.mpg.de}}

\maketitle

\newcommand{\TODO}[1]{{\color{red}{[TODO: #1]}}}
\newcommand{\phil}[1]{{\color[rgb]{0.3,0.7,0.3}{#1}}}
\newcommand{\zw}[1]{{\color[rgb]{0.7,0.3,0.7}{[ZW: #1]}}}
\newcommand{\jl}[1]{{\color[rgb]{0.7,0.7,0.3}{[JL: #1]}}}
\newcommand{\para}[1]{\vspace{.05in}\noindent\textbf{#1}}
\def\ie{\emph{i.e.}}
\def\eg{\emph{e.g.}}
\def\etal{{\em et al.}}
\def\etc{{\em etc.}}

\ifx\allfiles\undefined
\documentclass[letterpaper]{article}
\begin{document}
\else
\chapter{abstraction}
\fi

\begin{abstract}
This paper presents a new approach to boost a single-modality (LiDAR) 3D object detector by teaching it to simulate features and responses that follow a multi-modality (LiDAR-image) detector.
The approach needs LiDAR-image data only when training the single-modality detector, and once well-trained, it only needs LiDAR data at inference.
We design a novel framework to realize the approach:
response distillation to focus on the crucial response samples and avoid most background samples;
sparse-voxel distillation to learn voxel semantics and relations from the estimated crucial voxels;
a fine-grained voxel-to-point distillation to better attend to features of small and distant objects; and
instance distillation to further enhance the deep-feature consistency.
Experimental results on the nuScenes dataset show that our approach outperforms all SOTA LiDAR-only 3D detectors and even surpasses the baseline LiDAR-image detector on the key NDS metric, filling $\sim$72$\%$ mAP gap between the single- and multi-modality detectors.
\end{abstract}

\ifx\allfiles\undefined
\end{document}
\fi 
\ifx\allfiles\undefined
\documentclass[letterpaper]{article}
\begin{document}
\else
\chapter{introduction}
\fi
\section{Introduction}

State-of-the-art 3D object detectors,~\eg,~\cite{yin2021center,zheng2021se,shi2020pv,he2020structure,yang2019std}, widely adopt LiDAR-produced point clouds as the major input modality, since point clouds offer precise depth information and are robust to varying weather condition and illumination.
Yet, due to the laser-ray divergence, the sparsity of point clouds increases with distance.
So, there are only few points in small and distant objects, making it very hard to predict their object boundaries and semantic classes.

On the other hand, camera-produced images are a popular modality for monocular and stereo 3D object detection~\cite{luo2021m3dssd,reading2021categorical,chen2020dsgn,li2019stereo,ma2021delving}.
As images offer clear appearance and texture with dense pixels, image-based detectors can easily recognize the object boundaries and classify even small and distant objects.
Yet, images have no depth information and the visibility of objects depends on the environment conditions,~\eg, lighting,
so image-based detectors usually cannot be as accurate and robust as the LiDAR-based ones.

Some recent works~\cite{liang2019multi,wang2019frustum,vora2020pointpainting,yoo20203d,pang2020clocs} started to explore the fusion of 3D point clouds and 2D RGB images for improving the feature quality.
While higher precisions can often be attained, multi-modality detectors unavoidably sacrifice the inference efficiency for processing the extra modality.
Also, it is tedious to calibrate and synchronize different sensors, spatially and temporally, for high-quality data fusion.
Last, a breakdown of any modality sensor will cause a detector failure, thus reducing the system's fault tolerance.

\begin{figure}
\centering
\includegraphics[width=8cm]{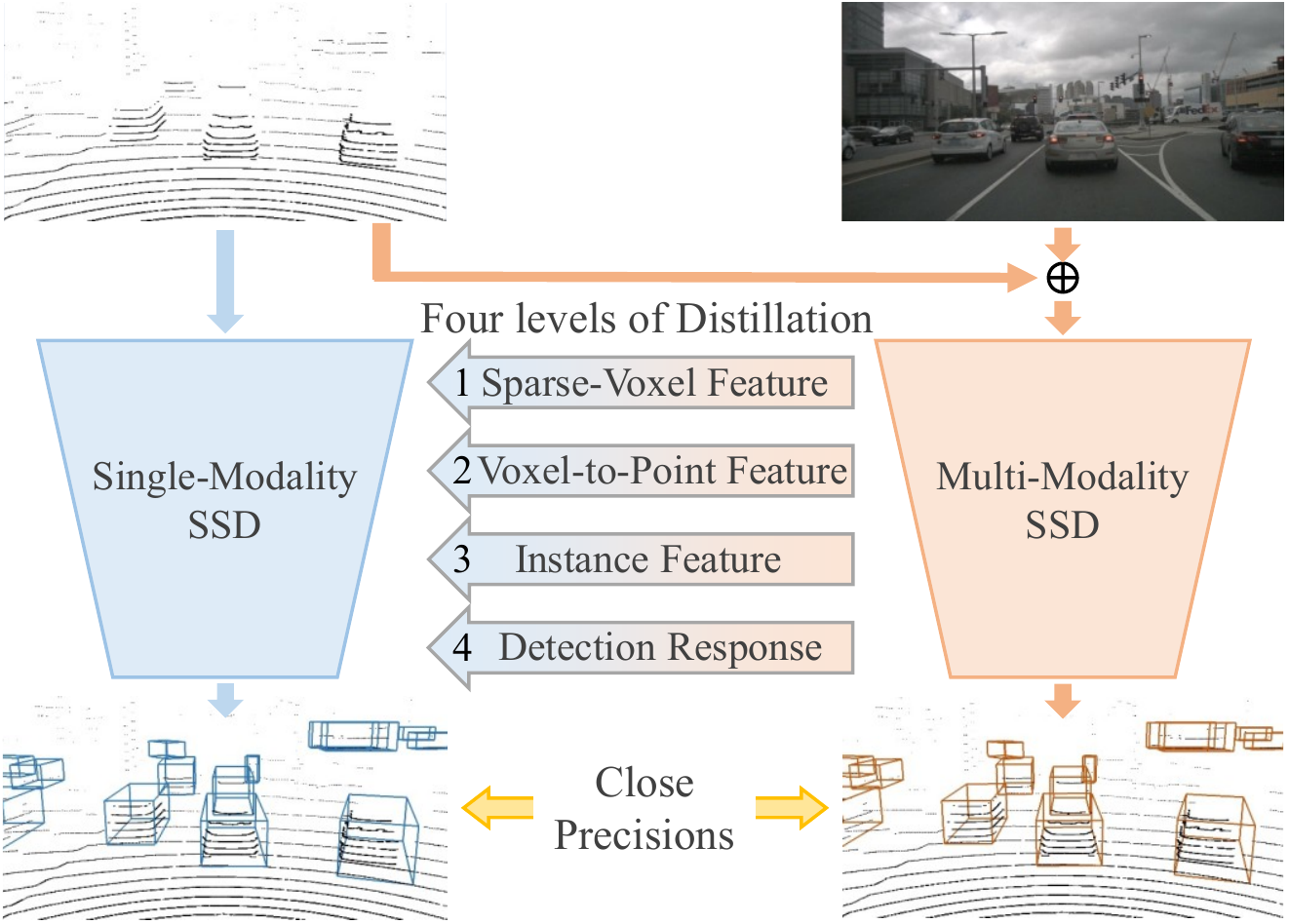}
\vspace*{-1.5mm}
\caption{
Overview of our Simulated Single-to-Multi-Modality Single-Stage 3D object Detector (S2M2-SSD) framework, by which we can train a single-modality SSD to learn from a multi-modality SSD and to achieve a high precision close to the multi-modality SSD but with only single-modality input at inference.
}
\label{fig:cover}
\vspace*{-4mm}
\end{figure}

\begin{figure*}
\centering
\includegraphics[width=0.992\textwidth]{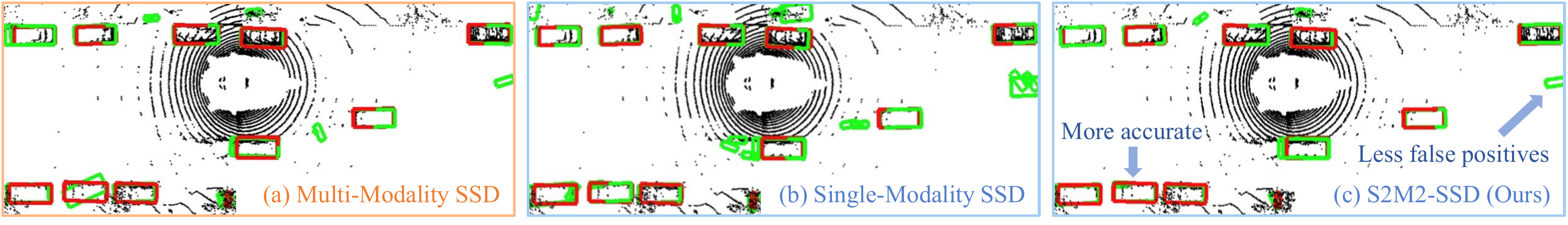}
\vspace*{-1.75mm}
\caption{
Even with only LiDAR input, our S2M2-SSD (c) is able to predict accurate bounding boxes (green) that well match the ground truths (red).
Its precision is close to that of the multi-modality SSD (a) and outperforms the SOTA LiDAR-only SSD~\cite{nabati2021centerfusion} (b).
}
\label{fig:cover2}
\vspace*{-3mm}
\end{figure*}

In this work, we propose {\em to teach a single-modality network to produce simulated multi-modality features and responses from only LiDAR input\/} by training the network to learn from 
a multi-modality LiDAR-image detector.
Our approach needs multi-modality data only when training the single-modality network, and once it is well-trained, it can detect 3D objects without image inputs.
This approach perfectly meets the sheer practical need of autonomous driving and boosts single-modality 3D object detection for
(i) {\em high efficiency\/}, since our approach needs to process only LiDAR data at inference;
(ii) {\em high precision\/}, since our network outperforms the SOTA LiDAR-only detectors; and
(iii) {\em high robustness\/}, since our approach is capable of simulating LiDAR-image features for detecting objects in varying lighting conditions, even at night time.

To realize the approach, the single-modality network has to effectively learn from the multi-modality network.
Yet, there are several technical challenges.
First, the vast 3D space is dominated by background samples, which hinder the transfer of foreground knowledge.
Second, 3D detectors involve massive points and/or voxels; naively distilling all pairs of voxel/point features from multi- to single-modality is
computationally infeasible.
Last, it remains challenging to effectively transfer knowledge for objects of various sizes and shapes, particularly for the small and distant ones.

We address the challenges by designing a novel Simulated Single-to-Multi-Modality Single-Stage 3D object Detector (S2M2-SSD) framework (see Figure~\ref{fig:cover}) to effectively train the single-modality network to learn from the multi-modality one.
This work has the following technical contributions.
First, we design the crucial response mining strategy and formulate the response distillation for focusing on the crucial responses while avoiding most background ones.
Second, we extend the strategy to voxels and formulate consistency constraints on voxel features and voxel relations to enhance the single-modality intermediate features.
Third, we formulate a fine-grained voxel-to-point distillation on the crucial foreground points for enhancing the features of objects with sparse points or of small sizes.
Last, we further correct the single-modality predictions by learning on the last-layer bird's eye view (BEV) features to improve the instance-level consistency in the deep-layer features.

The above techniques enable us to train and produce a single-modality network that takes only point clouds as input, yet capable of achieving a high performance close to a multi-modality LiDAR-image network; see Figure~\ref{fig:cover2}.
The evaluation on the nuScenes test set also shows that our S2M2-SSD outperforms all state-of-the-art single-modality 3D object detectors; our NDS metric even surpasses the multi-modality SSD and our mAP fills more than 70$\%$ of the gap between the single- and multi-modality SSDs.

\ifx\allfiles\undefined
\end{document}
\fi

\ifx\allfiles\undefined
\documentclass[letterpaper]{article}
\begin{document}
\else
\chapter{related_work}
\fi
\section{Related Work}

\begin{figure*}
\centering
\includegraphics[width=16.25cm]{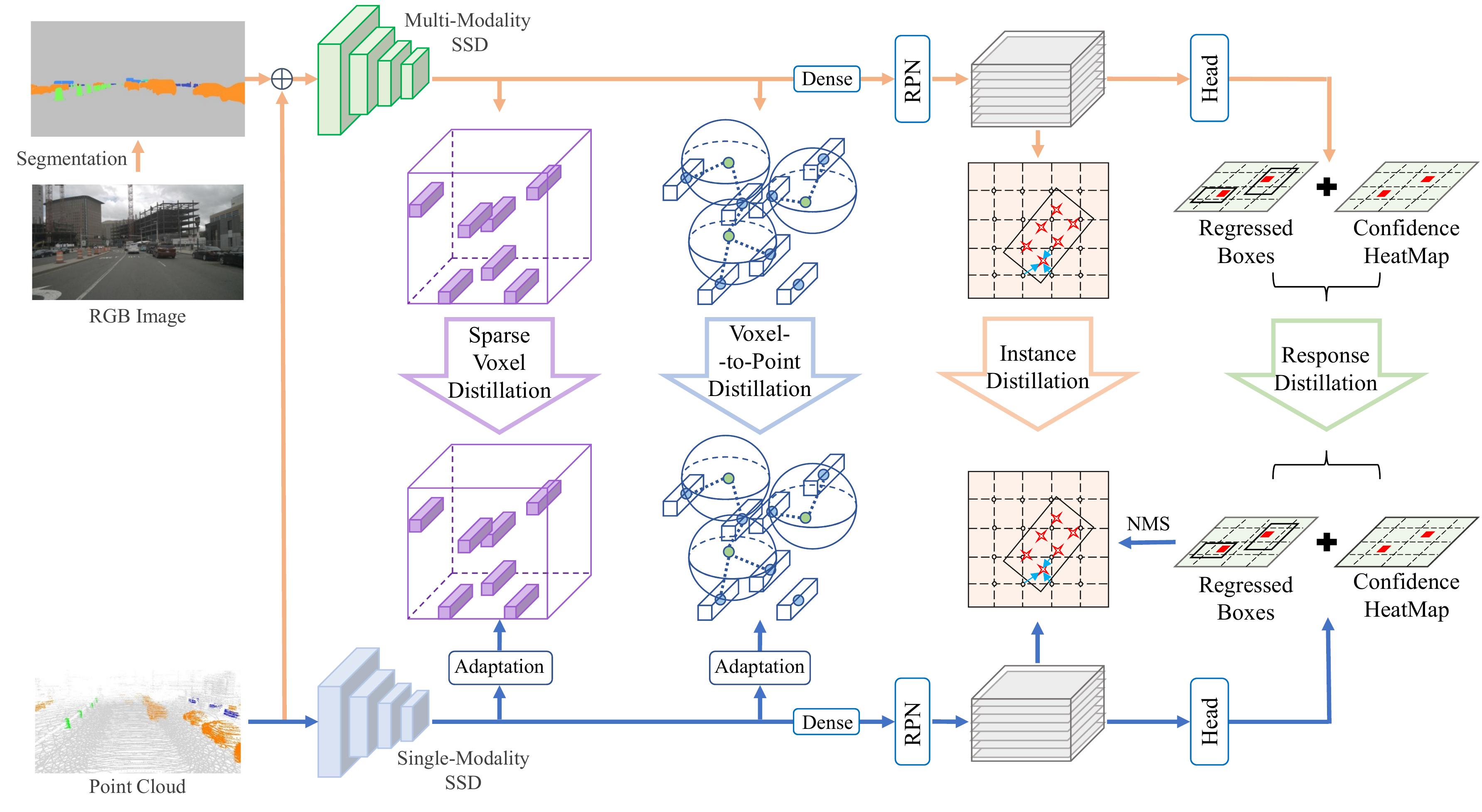}
\vspace*{-1.75mm}
\caption{
The pipeline of our S2M2-SSD framework.
The single-modality SSD (bottom) takes only a point cloud as input, whereas the multi-modality SSD (top) further takes a segmented image.
The training has two phases: first, we pre-train the multi-modality SSD (orange arrows).
Then, we train the single-modality SSD (orange and blue arrows) to learn to effectively produce features and responses comparable with the pre-trained multi-modality SSD by designing four levels of knowledge distillation in the framework: response distillation (Section \ref{sec:3.2}), sparse-voxel distillation (Section \ref{sec:3.3}), voxel-to-point distillation (Section \ref{sec:3.4}), and instance distillation (Section \ref{sec:3.5}).
During the testing (only the blue arrows), we only need a point cloud as the input of the well-trained single-modality SSD for the object detection.
}
\label{pipeline}
\vspace*{-2.25mm}
\end{figure*}

Mainstream 3D object detectors can generally be divided into two categories:
(i) single-modality detectors with either point clouds~\cite{zhou2018voxelnet,lang2019pointpillars,yan2018second,liu2020tanet,sheng2021improving,mao2021voxel,mao2021pyramid,fan2021rangedet} or RGB images~\cite{luo2021m3dssd,reading2021categorical,chen2020dsgn,li2019stereo,ma2021delving} as input,
and (ii) multi-modality detectors~\cite{chen2017multi,qi2018frustum,liang2019multi} with both point clouds and RGB images as input.
Multi-modality detectors often have higher precisions benefited from the complementarity of point clouds and RGB images, while single-modality detectors usually have higher efficiency due to less computation overhead.

Among the single-modality detectors, the LiDAR-only two-stage detectors~\cite{shi2019pointrcnn,shi2020pv,shi2020points} focus on enhancing the region-proposal-aligned features to boost the precision.
Recently, LiDAR-only single-stage detectors~\cite{yang20203dssd,he2020structure,zheng2020cia,zheng2021se,yin2021center} gradually surpass the two-stage ones with higher precisions.
SE-SSD~\cite{zheng2021se} employs a self-ensembling framework to exploit hard and soft targets for model optimization.
CenterPoint~\cite{yin2021center} regresses a confidence heatmap for anchor-free 3D detection.
Besides, some recent monocular/stereo 3D detectors~\cite{liu2021autoshape,shi2021geometry,luo2021m3dssd,reading2021categorical,chen2020dsgn,ma2021delving} use only RGB images as input and obtain significant improvements, yet their performance is still lower than those of the LiDAR-only detectors.

Among the multi-modality detectors, the fusion of image and point cloud is the most popular.
F-PointNet~\cite{qi2018frustum} projects 2D region proposals detected on RGB images to 3D frustums for filtering point clouds for 3D detection.
3D-CVF~\cite{yoo20203d} fuses semantics from multi-view images adaptively with point features.
CLOCs PVCas~\cite{pang2020clocs} refines the predicted confidence with features from images and point clouds.
PointPainting~\cite{vora2020pointpainting} combines the segmentation scores of images with LiDAR points as input.
PointAugmenting~\cite{wang2021pointaugmenting} performs late fusion between point and image segmentation features.
So far, only a few studies~\cite{nabati2021centerfusion,nobis2019deep,chadwick2019distant} attempt to fuse radar and image data, yet the performance still cannot surpass that of the LiDAR-image methods.

Unlike previous works, we fuse LiDAR and image data only in training and design four levels of knowledge distillation to effectively train a single-stage 3D object detector to learn to produce/simulate LiDAR-image features and responses from LiDAR-only data.
Knowledge distillation~\cite{hinton2015distilling} is first proposed for model compression and is widely applied in image classification~\cite{yim2017gift,heo2019comprehensive,tung2019similarity}.
Recently, a few 2D detectors~\cite{chen2021distilling,guo2021distilling,dai2021general,qi2021multi} explore decoupling or enriching the BEV features for knowledge distillation.
To the best of our knowledge, this work is the first attempt in 3D object detection on distilling knowledge from multi- to single-modality, and we are able to train a single-modality SSD whose performance is close to a multi-modality one.

\ifx\allfiles\undefined
\end{document}
\fi 
\ifx\allfiles\undefined
\documentclass[letterpaper]{article}
\begin{document}
\else
\chapter{framework}
\fi

\section{Simulated Single-to-Multi-Modality SSD}

\subsection{Overall Framework}
\label{sec:3.1}
Figure~\ref{pipeline} shows the pipeline of our S2M2-SSD framework.
We pre-train a multi-modality SSD (top) on point clouds and segmented images, following~\cite{vora2020pointpainting}, then train a single-modality SSD (bottom) only on point clouds.
To effectively train the single-modality SSD to produce features and responses comparable with those of the multi-modality SSD, we design four levels of knowledge distillation:
\begin{itemize}
\vspace*{-1mm}
\item
Response distillation (Section \ref{sec:3.2}) exploits the knowledge in multi-modality responses to correct the single-modality responses based on the crucial response mining we designed for focusing the distillation on the responses that are crucial for precision calculation.
\vspace*{-1.25mm}
\item
Sparse-voxel distillation (Section \ref{sec:3.3}) extends the mining strategy from responses to voxels, with consistency constraints formulated on voxel features and relations to distill semantics and relation knowledge in crucial voxels from multi- to single-modality SSD.
\vspace*{-1.25mm}
\item
Voxel-to-point distillation (Section \ref{sec:3.4}) aims to simulate fine-grained features for objects with sparse points or of small sizes, by transforming coarse-grained voxel features to fine-grained point features then distilling the fined-grained features in a point-wise manner.
\item
Instance distillation (Section \ref{sec:3.5}) helps to correct the single-modality predictions by learning the deep-layer BEV features in the NMS-filtered bounding boxes.
\end{itemize}

\vspace*{-1.25mm}
Note also that both our single- and multi-modality SSDs adopt the open-source SOTA CenterPoint~\cite{yin2021center} as backbone.

\subsection{Response Distillation}
\label{sec:3.2}
One major challenge to transfer knowledge from multi- to single-modality responses is the {\em imbalance between foreground and background samples\/}, as background responses easily dominate the distillation.
Hence, many 2D detectors either perform an indiscriminate distillation only on the positive responses~\cite{valverde2021there,guo2021distilling,sun2020distilling} or simply ignore the response distillation~\cite{qi2021multi,chen2021distilling,liu20213d,wang2019distilling} due to the sample imbalance issue.
In 3D detection, such issue is even more severe than the 2D cases,
as objects are much more sparse in the 3D space.

\begin{figure}
\centering
\includegraphics[width=0.99\columnwidth]{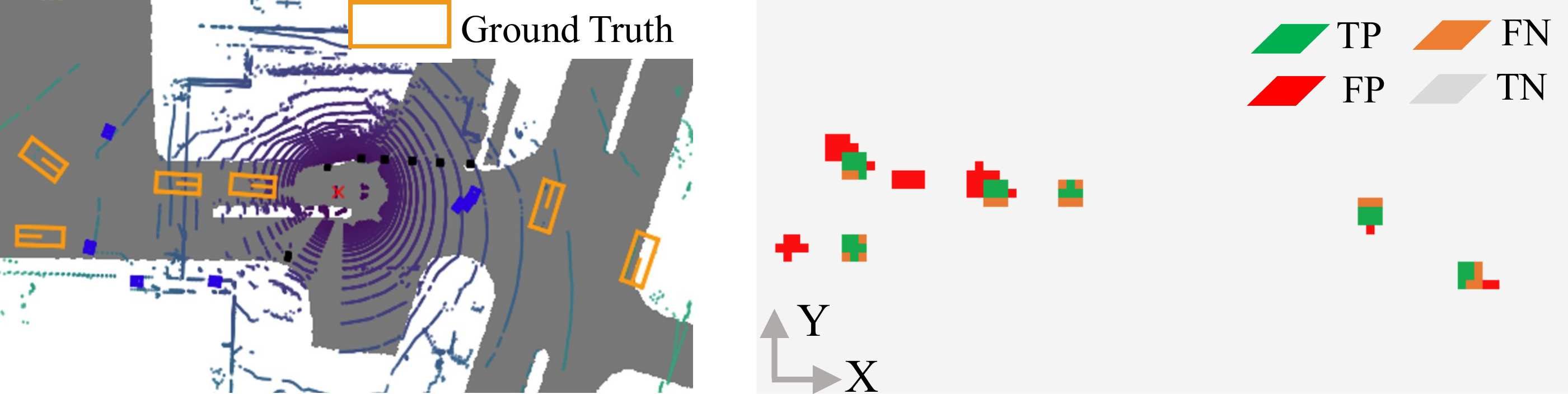}
\caption{
Left: a sample point cloud in bird's eye view (BEV) with the associated ground-truth bounding boxes (in orange).
Right: our estimated crucial responses for response distillation.
}
\label{fig:rsp_real}
\end{figure}

In our framework, we design the {\em crucial response mining\/} strategy to estimate the responses that are crucial for the detection precision of the single-modality SSD (subscript $s$),~\ie, true positives ($TP_{s}$), false positives ($FP_{s}$), and false negatives ($FN_{s}$).
We ignore true negatives, as they are very likely background samples.
For time efficiency, we find the crucial responses by comparing the confidence heatmap predicted by the single-modality SSD (denoted by $h_{s}$) and the one obtained from the ground truth (denoted by $h_{g}$).
The mechanism mimics how the IoU or the center distance metric measures the sample reliability:
\vspace*{-1mm}
\begin{equation}
\label{fpfn}
    \begin{split}
        & TP_{s} = (h_{s} > \tau) \; \& \;  (h_{g} >  \tau)  \\
        & FP_{s} = (h_{s} > \tau) \; \& \; (h_{g} <  \tau) \\
        & FN_{s} = (h_{s} < \tau) \; \& \; (h_{g} >  \tau)  \\
        \text{and} \ \
        & h_{d} = \max\big( h^{1}_{d}, \cdots, h^{K}_{d} \big), d \in \{s, g\}, \\
    \end{split}
\end{equation}

\vspace*{-1mm}
\noindent
where
$h^{i}_{d}$ is the confidence heatmap of the $i$-th class's bounding boxes from single-modality SSD or ground truth;
$K$ is the number of object classes in the detection head; and $\tau$ is a threshold.
By focusing the distillation on the crucial responses, we can avoid most background samples and make effective the response distillation;
see,~\eg, Figure~\ref{fig:rsp_real}.

With the estimated crucial responses, we can then formulate a weighted classification response distillation from the multi- to single-modality SSD by imposing a discriminative attention on different response classes:
\begin{equation}\label{rsploss1}
    \begin{split}
        & \mathcal L^r_{cls} = \frac{w^{r}_1}{|TP_s|}\sum_{i\in TP_s} \mathcal{L}^{r}_{\delta_h,i} + \frac{w^{r}_2}{|FP_s \cup FN_s|}\hspace*{-0.0mm}\sum_{i\in FP_s \cup FN_s} \hspace*{-4.5mm}\mathcal{L}^{r}_{\delta_h,i} \\
        &\text{and} \ \
        \mathcal{L}^{r}_{\delta_{h},i}=\mathcal{L}_{sml1}|h_{i,s}-h_{i,m}|, \
    \end{split}
\end{equation}
where $h_{m}$ is the multi-modality confidence heatmap corresponding to $h_{s}$;
$h_{i,m}$ is the $i$-th sample of $h_{m}$;
$\mathcal{L}_{sml1}$ is the smooth-$L_1$ loss;
and $w^{r}_1$ and $w^{r}_2$ are weights; we empirically set a larger $w^{r}_2$ to focus more on the false predictions.

Last, we perform regression distillation only on the true positive and false negative responses since they have associated ground-truth objects, while ignoring the false positive ones.
As image features offer clear object boundaries,
the multi-modality SSD may predict bounding boxes more accurately in some attributes,~\eg, size.  Hence, we impose differentiated weights on different bounding box attributes in the regression distillation:
\begin{equation}\label{rsploss2}
    \begin{split}
        &\mathcal L^r_{loc} = \frac{1}{|TP_s \cup FN_s|}\sum_{i\in TP_s \cup FN_s} \sum_{e} w^{r}_{e} \mathcal{L}^{r}_{\delta_e,i} \\
        &\text{and} \ \
        \mathcal{L}^{r}_{\delta_{e},i}=\mathcal{L}_{sml1}|e_{i,s}-e_{i,m}|, \\
    \end{split}
\end{equation}
where attribute $e$ $\in\{x, y, z, w, l, h, v_x, v_y, sin\theta, cos\theta \}$ and $\{w^{r}_e\}$ are attribute-wise weights.
Combining Eqs~\eqref{rsploss1} and~\eqref{rsploss2}, we obtain the overall response distillation loss:
\begin{equation}\label{over_cons}
        \mathcal L_{rsp} = \mathcal L^r_{cls} + \mathcal L^r_{loc} . \
\end{equation}

\begin{figure}
\centering
\includegraphics[width=0.99\columnwidth]{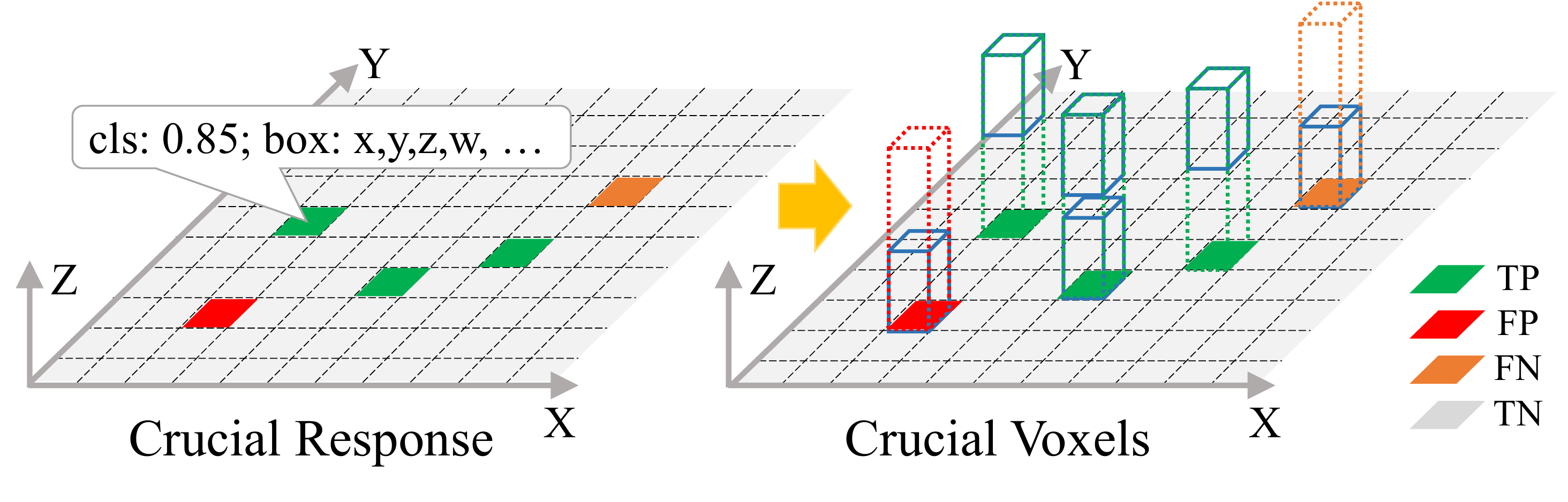}
\caption{
Illustration of crucial voxel mining.
We extend each crucial response sample into a pillar and take the active voxels in each pillar as crucial voxels to effectively avoid background voxels.
}
\label{fig:crucial}
\end{figure}

\subsection{Sparse-Voxel Distillation}
\label{sec:3.3}
Next, we design the sparse-voxel distillation to further enhance the single-modality SSD by exploring the voxel features in the last sparse convolution layer.
This layer has rich semantics and keeps the original 3D spatial information.
Compared to response distillation, sparse-voxel distillation can better {\em promote the consistency between the high-dimensional features\/} in single- and multi-modality SSDs.

To distill intermediate features, one may naively impose consistency constraints on all pairs of student and teacher features, as in 2D detection methods~\cite{guo2021distilling,qi2021multi,chen2021distilling,chen2017learning}.
Yet, 3D detection involves much more sparse target objects, so background features dominate the distillation and hinder the foreground knowledge transfer.
Also, calculating massive voxels is very time- and resource-consuming.
Hence, to consider the computing efficiency and avoid the background features, we leverage the crucial responses estimated from response distillation
to find active nonempty {\em crucial voxels\/} in pillars extended from the crucial response samples; see Figure~\ref{fig:crucial}.
As intermediate voxel features are shared by six detection heads for different class groups, we concatenate voxels of ${TP}_s$, ${FN}_s$, and ${FP}_s$ from different detection heads as ${TP}^{v}_s$, ${FN}^{v}_s$, and ${FP}^{v}_s$, respectively.

\begin{figure}
\centering
\includegraphics[width=0.99\columnwidth]{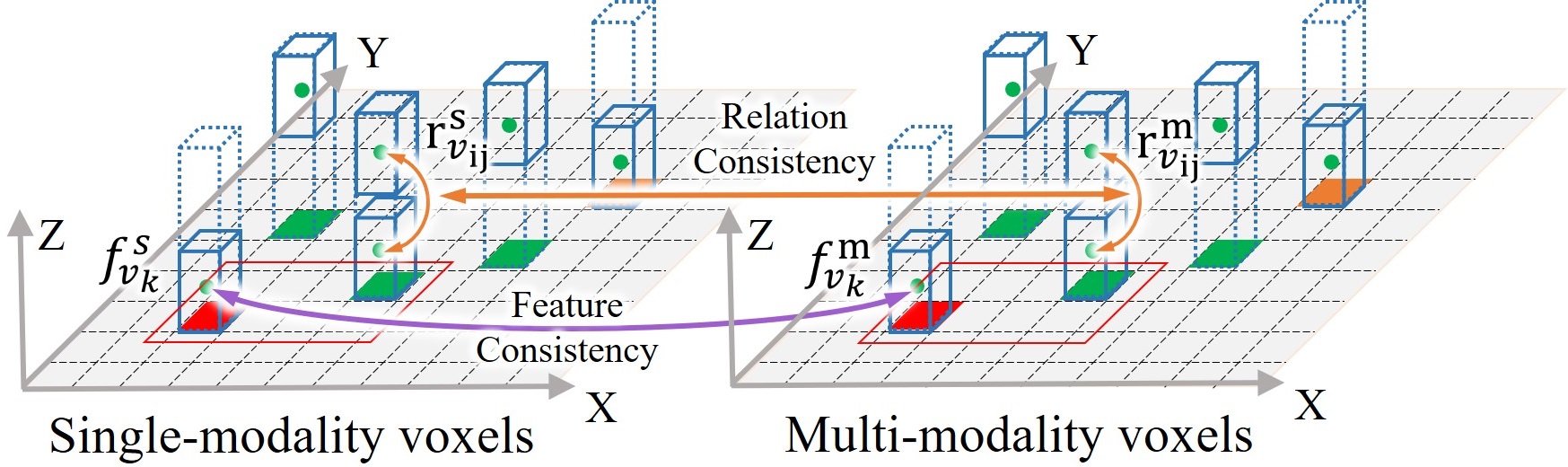}
\caption{
Our sparse-voxel distillation encourages voxel-feature consistency between the single- and multi-modality SSDs in two aspects: voxel features (purple) and voxel relations (orange).
}
\label{fig:consrelloss}
\end{figure}

With the crucial voxels, we then formulate a weighted voxel-wise consistency constraint between the single- and multi-modality SSDs (see the purple arrow in Figure~\ref{fig:consrelloss}) with differentiated weights imposed on different voxel classes:
\begin{equation}\label{fealoss1}
    \begin{split}
        &\mathcal L^{v}_{fea} = \frac{w^{v}_1}{|{TP}^{v}_s|}\sum_{i	\in {TP}^{v}_s} \mathcal{L}^{v}_{\delta_{f_i}}
        \hspace*{-1mm}
        +
        \hspace*{-0.5mm}
        \frac{w^{v}_2}{|{FP}^{v}_s \hspace*{-0.5mm} \cup \hspace*{-0.5mm} {FN}^{v}_s|}
        \sum_{i\in {FP}^{v}_s \cup {FN}^{v}_s}
        \hspace*{-4.5mm}
        \mathcal{L}^{v}_{\delta_{f_i}} \\
        &\text{and} \ \
        \mathcal{L}^{v}_{\delta_{f_i}}=\frac{1}{C} \mathcal L_{sml1}|f^{s}_{v_i} - f^{m}_{v_i}|,
    \end{split}
\end{equation}
where $f^{s}_{v_i}$
and $f^{m}_{v_i}$
are $C$-dimensional single- and multi-modality features, respectively, of the $i$-th crucial voxel;
$w^{v}_1$ and $w^{v}_2$ are weights; a larger $w^{v}_2$ is empirically set to focus more on voxel features associated with false predictions.

High-dimensional voxel features contain rich semantics and also voxel relations,~\eg,
different parts of the same object can have very different features, whereas similar parts in different objects can have similar features.
By exploiting the relation knowledge between voxels in multi-modality SSD, we can train the single-modality voxel features to be more discriminative; see the orange arrows in Figure~\ref{fig:consrelloss}.
So, we formulate the voxel relation consistency loss as
\begin{equation}\label{fealoss2}
    \begin{split}
        &\mathcal L^{v}_{rel} = \frac{1}{|V_s|^2}\sum_{i \in V_s}\sum_{j\in V_s}||r^{s}_{v_{ij}}-r^{m}_{v_{ij}}||^{2}_{2} \\
        & \text{and} \ \
        r^{d}_{v_{ij}} = \frac{f^{d}_{v_i}\cdot f^{d}_{v_j}}{{||f^{d}_{v_i}||}_{2}\cdot{||f^{d}_{v_j}||}_{2}}, d \in \{s, m\},
    \end{split}
\end{equation}
where
the crucial voxel set $V_s$\;=\;${TP}^v_s \cup {FP}^v_s \cup {FN}^v_s$,
and $r^{d}_{v_{ij}}$ is the cosine similarity between voxel features $f^{d}_{v_i}$ and $f^{d}_{v_j}$.
Hence, our sparse-voxel distillation loss is
\begin{equation}\label{over_cons}
        \mathcal L_{vxl} = \mathcal L^{v}_{fea} + \mathcal L^{v}_{rel} . \
\end{equation}

\subsection{Voxel-to-Point Distillation}
\label{sec:3.4}
For objects with sparse points or of small sizes, the low-resolution coarse-grained voxels in the last layer may not be able to capture their fine features, so sparse-voxel distillation may not be effective for these objects.
Alternatively, using high-resolution voxels in shallow layers is infeasible, due to computing inefficiency and insufficient semantics.

To circumvent the issue, we design the voxel-to-point feature distillation module for better handling
objects with sparse points or of small sizes.
Our key idea is to interpolate the
voxel features to raw point clouds and perform {\em fine-grained feature distillation on points\/}.
Yet, calculating all raw points is computationally infeasible.
So, we filter the points that are inside the ground-truth bounding boxes as {\em crucial foreground points\/} (denoted as $P$=$\{p_i\}_{i=1}^M$) for distillation.
Specifically,
denoting $p_{v_j}$ as the center coordinate of voxel $v_j$, we employ the feature propagation layer~\cite{qi2017pointnet++} to obtain the interpolated point feature at $p_i$ using
the inverse distance-weighted average to fuse nearby voxel features:
\begin{equation}\label{over_cons}
        f_{p_i} = \frac{\sum_{j=1}^{k} w_{ij} f_{v_j}} {\sum_{j=1}^{k}w_{ij}}, \;\; \text{where} \;\; w_{ij}=\frac{1}{||p_{i}-p_{v_j}||_2};
\end{equation}
$f_{p_i}$ denotes $p_i$'s point feature; and $k$ denotes the number of voxels in the neighborhood of $p_i$.
With $f_{p_i}$, we can then perform the point-wise feature distillation between single- and multi-modality SSDs with the consistency loss:
\begin{equation}\label{vtploss1}
    \begin{split}
        &\mathcal L^{p}_{fea} = \frac{w^p_f}{|P|}\sum_{i\in P} \frac{1}{C} \mathcal L_{sml1}|f^{s}_{p_{i}} - f^{m}_{p_{i}}|, \\
    \end{split}
\end{equation}
where $w^p_f$ is weight and $C$ is the number of channels in $f_{p_i}$.
Further, we
can exploit the rich relation knowledge among points by
formulating the point relation distillation loss:
\begin{equation}\label{vtploss2}
    \begin{split}
        &\mathcal L^{p}_{rel} = \frac{1}{|P'|^2}\sum_{i\in P'}\sum_{j\in P'}||r^{s}_{p_{ij}}-r^{m}_{p_{ij}}||^{2}_{2} \\
        & \text{and} \ \
        r^{d}_{p_{ij}} = \frac{f^{d}_{p_i}\cdot f^{d}_{p_j}}{{||f^{d}_{p_i}||}_{2}\cdot{||f^{d}_{p_j}||}_{2}}, d \in \{s, m\},
    \end{split}
\end{equation}
where
$r^{d}_{p_{ij}}$ denotes the cosine similarity between point features $f^d_{p_i}$ and $f^d_{p_j}$; and
$P'$ is the set of $\min(M, 4500)$ points randomly selected from $P$ to avoid excessive GPU computation.
Combining Eqs.~\eqref{vtploss1} and~\eqref{vtploss2}, we formulate the voxel-to-point distillation loss as
\begin{equation}\label{over_cons}
        \mathcal L_{pts} = \mathcal L^{p}_{fea} + \mathcal L^{p}_{rel}.
\end{equation}

\begin{table*}[t]
\centering
\caption{Comparison with SOTA LiDAR-only detectors on the nuScenes test set.
Our S2M2-SSD attains the {\em highest NDS and mAP\/}, as well as {\em highest AP consistently for all ten object classes\/}.
The percentages in () mean the proportions of S2M2-SSD's gains on the single-modality SSD relative to the single- and multi-modality metric gap.
`*' means the SSD is built on our improved version of CenterPoint.}
 \label{table1}
 \vspace*{-2mm}
\resizebox{\linewidth}{!}{
\begin{tabular}{l@{\ \ \ }c@{\ \ \ }c@{\ \ \ }c@{\ \ \ }c@{\ \ \ }c@{\ \ \ }c@{\ \ \ }c@{\ \ \ }c@{\ \ \ }c@{\ \ \ }c@{\ \ \ }c@{\ \ \ }c@{\ \ \ }c}
  \toprule
  Method                           & Modality  & NDS & mAP  & Car & Truck & Bus & Trailer & CV & Ped & Motor & Bicycle & TC & Barrier \\
 \cmidrule(r){1-1}
 \cmidrule(r){2-2}
 \cmidrule(r){3-4}
 \cmidrule(){5-14}
  WYSIWYG~\cite{hu2019exploiting}           & LiDAR &41.9 &35.0 &79.1 &30.4 &46.6 &40.1 &7.1 &65.0 &18.2 &0.1 &28.8 &34.7 \\
  PointPillars~\cite{lang2019pointpillars}  & LiDAR &45.3 &30.5 &68.4 &23.0 &28.2 &23.4 &4.1 &59.7 &27.4 &1.1 &30.8 &38.9  \\
  PointPainting \cite{vora2020pointpainting}& LiDAR &58.1 &46.4 &77.9 &35.8 &36.2 &37.3 &15.8 &73.3 &41.5 &24.1 &62.4 &60.2  \\
  CVCNet \cite{chen2020view}                & LiDAR &64.4 &55.3 &82.7 &46.1 &46.6 &49.4 &22.6 &79.8 &59.1 &31.4 &65.6 &69.6  \\
  PMPNet \cite{yin2020Lidarbased}           & LiDAR &53.1 &45.4 &79.7 &33.6 &47.1 &43.1 &18.1 &76.5 &40.7 &7.9 &58.8 &48.8  \\
  SSN \cite{zhu2020ssn}                     & LiDAR &58.1 &46.4 &80.7 &37.5 &39.9 &43.9 &14.6 &72.3 &43.7 &20.1 &54.2 &56.3  \\
  CBGS \cite{zhu2019classbalanced}          & LiDAR &63.3 &52.8 &81.1 &48.5 &54.9 &42.9 &10.5 &80.1 &51.5 &22.3 &70.9 &65.7 \\
  CenterPoint~\cite{yin2021center}          & LiDAR &65.5 &58.0 &84.6 &51.0 &60.2 &53.2 &17.5 &83.4 &53.7 &28.7 &76.7 &70.9\\
  \cmidrule(){1-14}
  Multi-modality SSD*                       & {\em LiDAR+RGB\/}  &{\em 69.1} &{\em 64.0} &{\em 86.2} &{\em 55.4} &{\em 65.6} &{\em 58.2} &{\em 28.3} &{\em 85.0} &{\em 65.1} &{\em 40.0} &{\em 79.8} &{\em 75.9} \\
  Single-modality SSD*                      & LiDAR &67.3 &60.1 &85.2 &51.9 &63.6 &55.9 &21.7 &83.1 &55.7 &33.1 &75.7 &74.7 \\
  S2M2-SSD (Ours)    &LiDAR &\bf 69.3 &\bf 62.9 &\bf 86.3 &\bf 56.0 &\bf 65.4 &\bf 59.8 &\bf 26.2 &\bf 84.6 &\bf 61.6 &\bf 36.4 &\bf 77.7 &\bf 75.1 \\
  \rowcolor{LightCyan}
  \textit{Improvement} & - &\textit{+2.0 (111$\%$)} &\textit{+2.8 (72$\%$)}  &\textit{+1.1} &\textit{+4.1} &\textit{+1.8} &\textit{+3.9} &\textit{+4.5} &\textit{+1.5} &\textit{+5.9} &\textit{+3.3} &\textit{+2.0} &\textit{+0.4}  \\

  \bottomrule
 \end{tabular}
 }
\end{table*}


\subsection{Instance Distillation}
\label{sec:3.5}
While intermediate 3D feature distillations greatly promote the knowledge transfer from multi- to single-modality SSD, the single-modality features may still deviate from the multi-modality ones, as the layer goes deeper.
Distillation on low-dimensional responses also cannot guarantee consistent high-dimensional deep features.
To this end, we design the instance distillation module {\em on the last-layer BEV features to promote the consistency of the deep features\/}, which have a direct impact on the object prediction.

This module first uses NMS to remove redundant bounding boxes predicted on the single-modality BEV features, then
performs the rotated RoI-grid pooling to crop the last-layer BEV features in each filtered bounding box.
In detail, we treat each filtered bounding box as an instance, fit a uniform grid ($G$=$5$$\times$$5$) in each bounding box, interpolate BEV features at each grid point,
and transfer knowledge from multi- (superscript m) to single-modality (superscript s) through the $5$$\times$$5$ interpolated features:
\begin{equation}\label{insloss}
    \begin{split}
        &\mathcal L_{ins} = \frac{w^I}{B}\sum_{i=1}^{B} \frac{1}{G}\sum_{j=1}^{G}\mathcal L_{sml1}|f^{s}_{I_{ij}} - f^{m}_{I_{ij}}|, \
    \end{split}
\end{equation}
where $B$ is the number of NMS-filtered bounding boxes;
$f_{I_{ij}}$ is the interpolated feature at the $j$-th grid point of the $i$-th instance;
and $w^I$ is a hyper-parameter.


\subsection{Overall Loss Function}
\label{sec:3.6}
We train the single-modality SSD end-to-end (while fixing the pre-trained multi-modality SSD; see Figure~\ref{pipeline}) with the following supervision losses and distillation losses:
\begin{equation}\label{totalloss}
  \mathcal{L}= \mathcal L_{cls} + \lambda \mathcal L_{reg} +  \mu( \mathcal L_{rsp} + \mathcal L_{vxl} +  \mathcal L_{pts} + \mathcal L_{ins}), \
\end{equation}
where $L_{cls}$ is classification loss and $L_{reg}$ is regression loss, following~\cite{yin2021center}; and $\lambda$ and $\mu$ are hyper-parameters.

\ifx\allfiles\undefined
\end{document}
\fi

\ifx\allfiles\undefined

\documentclass[letterpaper]{article}
\begin{document}
\else
\chapter{experiments}
\fi

\section{Experiments}

\subsection{Experimental Setup}

\paragraph{Dataset.}
We use nuScenes~\cite{caesar2020nuscenes}, a popular large-scale multimodal dataset, in our experiments.
The dataset comprises 1,000 driving sequences (700/150/150 for train/val/test), each 20 seconds long, and every ten frames are fully annotated with 3D object bounding boxes.
The LiDAR sensor scans at a frequency of 20 FPS, producing 400 frames per sequence and around 30k points per frame.
For each frame, the dataset also provides RGB images from all six cameras to realize a complete 360$^{\circ}$ coverage.
The nuScenes detection task involves ten classes of objects of various sizes and shapes, including 28k, 6k, and 6k samples for training, validating, and testing, respectively.
The goal of the task is to determine the following parameters of each bounding box: x, y, z, width, length, height, velocity, and yaw angle.

\vspace*{-3mm}
\paragraph{Metrics.}
We adopt the main official evaluation metrics of nuScenes,~\ie, the nuScenes detection score (NDS) and mean Average Precision (mAP), which are calculated by averaging over all the object classes.
NDS is calculated halfly based on mAP and halfly based on
the true-positive qualities,~\ie, translation, velocity, orientation, etc., whereas
mAP is calculated based on matching boxes by thresholding the 2D center distance (0.5m, 1m, 2m, 4m) on the ground.

\paragraph{Network and Training.}
We follow CenterPoint~\cite{yin2021center} to build our single- and multi-modality SSDs and improve it by more data augmentation.
The detection range is [-51.2m, 51.2m] for the X and Y axes and [-5m, 3m] for the Z axis, and the point clouds are discretized by a voxel size of [0.1m, 0.1m, 0.2m].
To obtain a high precision, we employ the open-source PointPainting~\cite{vora2020pointpainting} to pre-train the multi-modality SSD with both point clouds and RGB images.
Also, we use a linear layer with ReLU activation as the adaptation layer for sparse-voxel distillation, and a submanifold sparse convolution layer with ReLU activation as the adaptation layer for voxel-to-point distillation.
We set
$\tau$ as 0.1 (Eq.~\eqref{fpfn}),
$w^{r}_1$ as 1.0 and $w^{r}_2$ as 5.0 (Eq.~\eqref{rsploss1}),
$\{w^{l}_{e}\}$ as 0, 0, 0, 0.1, 0.1, 0.1, 0.1, 0.1, 0, and 0 (Eq.~\eqref{rsploss2}),
$w^{v}_1$ as 2.0 and $w^{v}_2$ as 8.0 (Eq.~\eqref{fealoss1}),
$w^p_{f}$ as 2.0 (Eq.~\eqref{vtploss1}),
$w^I$ as 8.0 (Eq.~\eqref{insloss}),
$\lambda$ as 0.25 and $\mu$ as 0.5 (Eq.~\eqref{totalloss}).

\subsection{Comparison with the State-of-the-Arts}
We evaluated our S2M2-SSD on the nuScenes test set by submitting the predicted results to the nuScenes server. Table~\ref{table1} reports the resulting NDS and mAP for comparison with the state-of-the-art LiDAR-only methods.
As shown in the table, our S2M2-SSD attains the highest NDS and mAP among all LiDAR-only detectors, with significant gains of +2.8 points on mAP and +2.0 points on NDS over the single-modality SSD.
Also, the mAP gain (+2.8) of our S2M2-SSD fills $\sim$72$\%$ mAP gap (3.9) between the single- and multi-modality SSDs, while the NDS of our S2M2-SSD even surpasses that of the multi-modality SSD.
We think that our higher NDS is due to the further improvement in the true-positive qualities based on the mAP gain.

Also, our S2M2-SSD attains consistent improvements for all ten object classes over the state-of-the-art methods.
Compared with the multi-modality SSD, our S2M2-SSD attains comparable or even higher APs on `car', `truck', `trailer', `bus', and `pedestrian', since larger objects can be easily benefited from all of our distillation modules.
On the other hand, on `construction vehicle', `motorbike', `bicycle', and `traffic cone', our method is able to largely narrow down the AP gap between the single- and multi-modality SSDs.
As for `barrier', these objects have high variations in shapes and sizes, so are very hard for consistent distillation.

Besides, since S2M2-SSD only needs point clouds as input, it has high efficiency at inference, compared to multi-modality detectors that need to process both LiDAR and image inputs; see the details in Section~\ref{sec:4.4}.
Also, S2M2-SSD is able to produce simulated multi-modality features, even at night time, manifesting its robustness to the environment's lighting condition.
Lastly, note that while there are some higher-precision LiDAR-only and LiDAR-image methods on the nuScenes leaderboard, they do not have peer-reviewed papers and public code for us to experiment our method with.
So, we resort to use the open-source CenterPoint and PointPainting as our framework backbones.

{
\begin{table}[t]
\small
\begin{center}
\caption{Evaluation on the nuScenes validation set with models trained on 30$\%$ training data, serving as ablation study baselines.}
\label{table2}
\vspace*{-2mm}
\begin{tabular}{lccc}
  \hline
  Method                                            & Modality  & NDS & mAP  \\
  \hline
  Multi-modality SSD                                & LiDAR+RGB & {\em 55.1} & {\em 48.0} \\
  Single-modality SSD                               & LiDAR     & 51.0 & 42.0 \\
  S2M2-SSD                                          & LiDAR     & \bf55.6 & \bf46.2 \\
  \rowcolor{LightCyan}
  \textit{Improvement}                              & -         & \textit{+4.6} & \textit{+4.2} \\
  \hline
 \end{tabular}
\end{center}
\vspace{-3mm}
\small
\end{table}
}

\begin{table}[t]
\centering
\caption{Ablation study on our proposed modules: ``response'', ``voxel'', ``point'', and ``instance'' denote the response, sparse-voxel, voxel-to-point, and instance distillation modules, respectively.
All results are based on model training on 30$\%$ train set, and the NDS and mAP are reported on the nuScenes val split.
 }
\vspace*{-2mm}
\resizebox{0.88\columnwidth}{!}{
\begin{tabular}{cccc|cc}
    \hline
      \multicolumn{1}{c}{response} &\multicolumn{1}{c}{voxel} &\multicolumn{1}{c}{point} &\multicolumn{1}{c|}{instance}
      &\multicolumn{1}{c}{ \multirow{1}{*}{NDS}} &\multicolumn{1}{c}{ \multirow{1}{*}{mAP}} \\
      \hline
                  &            &            &            & 51.0     & 42.0        \\
       \checkmark &            &            &            & 53.0     & 43.9       \\
                  & \checkmark &            &            & 53.8     & 44.2       \\
                  &            & \checkmark &            & 52.7     & 43.8     \\
                  &            &            & \checkmark & 52.6     & 43.2       \\
       \checkmark & \checkmark &            &            & 54.3     & 44.9        \\
       \checkmark & \checkmark & \checkmark &            & 55.0     & 45.6         \\
       \checkmark & \checkmark & \checkmark & \checkmark & \bf55.6  & \bf46.2  \\ \hline
\end{tabular}
}
\label{table3}
\vspace{-1mm}
\end{table}

\subsection{Ablation Study}
For efficiency, we follow~\cite{wang2021pointaugmentingS,shi2021pv} to conduct all ablation studies by training on 30$\%$ train samples and evaluating on the complete val split.
Table~\ref{table2} shows the NDS and mAP of our S2M2-SSD and also the multi- and single-modality SSDs, which serve as base results in the ablation studies.
Table~\ref{table3} shows the effects of each module in S2M2-SSD.

{
\begin{table}[t]
\centering
\caption{Ablation study on response distillation, in which we show the effect of distilling different kinds of (crucial) responses, as well as the effect of regression distillation (denoted by ``loc'').}
\label{table4}
\begin{center}
\vspace{-5mm}
\resizebox{0.85\columnwidth}{!}{
\begin{tabular}{c|cc}
  \hline
  Distilled responses                        & NDS   & mAP   \\
  \hline
  baseline                                   & 51.0  & 42.0 \\
  all                                        & 51.4  & 42.3 \\
  true positives                             & 52.1  & 43.0 \\
  + false positives                          & 51.6  & 42.7 \\
  + false negatives                          & 51.8  & 42.6 \\
  + false positives \& negatives             & 52.8  & 43.7 \\
  + false positives \& negatives + loc       & \bf53.0  & \bf43.9 \\
  \hline
 \end{tabular}
}
\end{center}
\vspace*{-5mm}
\end{table}
}

\vspace*{-3mm}
\paragraph{Effect of response distillation.}
As the first two rows in Table~\ref{table3} show, our response distillation module boosts the NDS by 2.0 points and mAP by 1.9 points, showing that our approach can effectively improve the single-modality responses for higher consistency with the multi-modality ones.
Also, this module consumes much less computation than others, yet delivering impressive improvements.

Table~\ref{table4} shows further ablation studies on our crucial response mining strategy.
From the table, we can see that performing response distillation on all samples only boosts NDS by 0.4 and mAP by 0.3, while performing response distillation only on the true positives already improves NDS by 1.1 and mAP by 1.0.
These results show that the dominated background samples seriously hinder the transfer of knowledge for the foreground samples.
On the other hand, adding either false positives or false negatives reduces the performance, as doing so may significantly change the confidence relations between the distilled responses and the undistilled crucial responses.
When distilling all crucial responses, both NDS and mAP can further be improved by 0.7 (comparing 3rd and 6th rows).
Lastly, the regression distillation further improves both metrics by 0.2.

\vspace*{-3mm}
\paragraph{Effect of sparse-voxel distillation.}
Comparing the first and third rows in Table~\ref{table3} shows that our sparse-voxel distillation improves the NDS by 2.8 points and mAP by 2.2 points, which are the largest single-module improvements compared with others.
These results manifest the necessity of enhancing the high-dimensional intermediate features.
Also, comparing the second and sixth rows in Table~\ref{table3} shows that this module further increases the NDS by 1.3 points and mAP by 1.0 points on top of the response distillation, showing the complementary strengths of the two modules.

Table~\ref{table5} shows further ablation results conducted on the crucial voxel mining strategy and the consistency losses on voxel features and voxel relations.
By comparing the results between the 2nd-4th and 5th-7th rows, we can see that although our crucial voxel mining strategy keeps only 984 crucial voxels out of all the 7718 active voxels on average per 3D scene, it reduces the number of FLOPs significantly from 9.15$\times10^{10}$ to 1.49$\times10^{9}$, and boosts the precisions (both NDS and mAP) obviously, as it helps avoid the background voxels and overfitting them.
Note that the FLOPs for the relation loss has a time complexity of O($V^2$), where $V$ is the number of voxels.
Also, we study the effects of each item in the consistency losses; from 2nd-4th and 5th-7th rows, we can see that the voxel-feature loss contributes most, since it promotes direct knowledge transfer between the coarse-grained voxels.
Further, the voxel-relation loss can effectively boost the NDS (+0.7) more than mAP (+0.1), since it improves the true-positive metrics more by making the features more discriminative.

{
\begin{table}[t]
\centering
\caption{Ablation study on sparse-voxel distillation: ``filter'' means the crucial voxel mining; ``cons'' and ``rel'' denote the consistency losses on voxel features and relations, respectively; ``Voxels'' means the average number of voxels for distillation per point cloud; and ``FLOPs'' is calculated based on Eqs.~\eqref{fealoss1} and~\eqref{fealoss2}.}
\vspace{-5mm}
\begin{center}
\resizebox{0.98\linewidth}{!}{
\begin{tabular}{c|c|c|cc}
  \hline
  Method                                & Voxels  & FLOPs                   & NDS  & mAP  \\
  \hline
  baseline                              & -       & -                       & 51.0  & 42.0 \\
  \hline
  cons w/o filter                       &         & 1.98$\times10^{6}$      & 52.7 & 43.2 \\
  rel w/o filter                        & 7718    & 9.15$\times10^{10}$     & 52.0 & 42.4 \\
  cons\&rel w/o filter                  &         & 9.15$\times10^{10}$     & 53.3  & 43.4 \\
  \hline
  cons w/  filter                       &         & 2.52$\times10^{5}$      & 53.1 & 44.1 \\
  rel w/ filter                         & 984     & 1.49$\times10^{9}$      & 52.5 & 42.7 \\
  cons\&rel w/ filter                   &         & 1.49$\times10^{9}$      & \bf53.8 & \bf44.2 \\
  \hline
 \end{tabular}
}
\end{center}
\vspace{-5mm}
\label{table5}
\end{table}
}

{
\begin{table}[t]
\centering
\caption{Ablation study on voxel-to-point distillation, manifesting the effects of consistency losses on point features and relations, especially for objects with sparse points or of small sizes.}
\vspace{-5mm}
\begin{center}
\resizebox{0.999\linewidth}{!}{
\begin{tabular}{@{\hspace*{0.5mm}}c@{\hspace*{1.5mm}}|@{\hspace*{1.5mm}}c@{\hspace*{1.5mm}}c@{\hspace*{1.5mm}}c@{\hspace*{1.5mm}}c@{\hspace*{1.5mm}}c@{\hspace*{1.5mm}}|@{\hspace*{1.5mm}}c@{\hspace*{1.5mm}}c@{\hspace*{0.5mm}}}
  \hline
  Method                          & ped    & motor  & bicycle & TC   & barrier & NDS & mAP  \\
  \hline
  baseline                        & 72.2   & 38.6   & 11.7    & 49.6 & 49.5    & 51.0 & 42.0 \\
  cons loss                       & 75.6   & 38.0   & 13.2    & 51.9 & 50.7    & 52.2 & 43.2 \\
  rel loss                        & 75.1   & 39.1   & 13.4    & 52.0 & 51.0    & 52.6 & 43.3 \\
  cons\&rel loss                  & \bf76.1   & \bf40.7   & \bf13.7    & \bf52.3 & \bf51.7  & \bf52.7   & \bf43.8 \\
  \hline
 \end{tabular}

}
\end{center}
\label{table6}
\vspace{-5mm}
\end{table}
}

\vspace*{-3mm}
\paragraph{Effect of voxel-to-point distillation.}
Comparing the first and forth rows in Table~\ref{table3} shows that our voxel-to-point distillation improves NDS by 1.7 and mAP by 1.8.
Comparing the sixth and seventh rows shows that it still increases both NDS and mAP by 0.7 on top of the previous two modules, showing the importance of simulating fine-grained features.

Table~\ref{table6} further shows the module's effects on objects with sparse points or of small sizes.
It can be seen that the `pedestrian' AP can be significantly improved by 3.9 points, and the APs of `motorbike', `bicycle', `TC', and `barrier' are also improved by around 2$\sim$3 points, validating our motivation of promoting the fine-grained feature consistency.
Also, we can see that the point-relation loss contributes more than the point-feature loss, opposite to the effects of sparse-voxel distillation, as the point-relation loss can help produce more discriminative single-modality features by exploiting and contrasting the point features.

{
\begin{table}[t]
\centering
\caption{Ablation study on instance distillation, in which ``gt boxes'', ``crucial-nms'', and ``all-nms'' denote the instance defined by the ground-truth boxes, NMS-filtered crucial responses, and all predicted bounding boxes filtered by NMS, respectively.}
\vspace{-5mm}
\begin{center}
\resizebox{0.98\linewidth}{!}{
\begin{tabular}{c|cccc}
  \hline
  Method                     & baseline  & gt boxes & crucial-nms & all-nms (ours) \\
  \hline
  NDS                        & 51.0      & 51.8     & 52.4        & \bf52.6 \\
  mAP                        & 42.0      & 42.5     & 42.7        & \bf43.2 \\
  \hline
 \end{tabular}
 }
\end{center}
\label{table7}
\vspace{-5mm}
\end{table}
}

{
\begin{table}[t]
\centering
\caption{Comparing the runtime (in millisecond) of our S2M2-SSD against the multi-modality SSD for processing each modality data, showing the high computational efficiency of our approach.}
\vspace{-5mm}
\begin{center}
\resizebox{0.85\linewidth}{!}{
\begin{tabular}{c|cc|c}
  \hline
  Method                     & image  & point cloud & total \\
  \hline
  Multi-modality SSD         & 425.5   & 107.7      & 533.2 \\
  S2M2-SSD (ours)            & -      & 81.9        & {\bf 81.9} \\
  \hline
 \end{tabular}
}
\end{center}
\vspace{-5mm}
\label{table8}
\end{table}
}

\vspace*{-3mm}
\paragraph{Effect of instance distillation.}
Comparing the first and fifth rows in Table~\ref{table3} shows that instance distillation improves the NDS by 1.6 and mAP by 1.2.
Also, instance distillation still improves both two metrics by 0.6 points (see the last two rows) over the previous modules, showing its effect in enhancing the deep features and correcting the single-modality predictions.
Table~\ref{table7} shows more ablation results conducted with different definitions of instances, including the ground truths, NMS-filtered crucial responses, and NMS-filtered bounding boxes predicted by the single-modality SSD.
We can see that our method attains better results than the other settings, since our focused deep features have direct association with the final detected boxes.

\subsection{Runtime Analysis}
\label{sec:4.4}
Table~\ref{table8} compares the runtime of our S2M2-SSD and the multi-modality SSD to show the high efficiency of our approach.
As shown in the table, S2M2-SSD only needs 81.9ms for detecting objects in a 3D point cloud, whereas the multi-modality SSD needs 425.5ms and 107.7ms to process the input image and detect objects in the point cloud, respectively.
Note that we ignore the time for calibration, synchronization, and data fusion, which are hard to be quantified.
All evaluations were done on an Intel Xeon Silver CPU and a TITAN Xp GPU with a batch size of four.

\ifx\allfiles\undefined
\end{document}
\fi 
\ifx\allfiles\undefined

\documentclass[letterpaper]{article}
\begin{document}
\else
\chapter{Conclusion}
\fi

\section{Conclusion}
We presented a novel framework capable of producing a single-stage 3D object detector with high precision, efficiency, and robustness, by simulating multi-modality (LiDAR-image) on single-modality (LiDAR) input.
Our S2M2-SSD framework consists of four levels of knowledge distillation:
response distillation to focus on the crucial responses and avoid most background samples;
sparse-voxel distillation to learn the voxel semantics and relation knowledge;
voxel-to-point distillation to attend to features of small and distant objects;
and instance distillation to promote deep-layer feature consistency.
Experimental results on nuScenes shows that our S2M2-SSD achieves SOTA LiDAR-only performance and surpasses the baseline multi-modality SSD on the key NDS metric, filling $\sim$72$\%$ mAP gap between the single- and multi-modality SSDs.

\ifx\allfiles\undefined
\end{document}
\fi

\vspace*{1mm}
\noindent
{\bf Acknowledgments.} \
This research is supported by project MMT-p2-21 of the Shun Hing Institute of Advanced Engineering, The Chinese University of Hong Kong.

{\small
\bibliographystyle{ieee_fullname}
\bibliography{egbib}
}

\end{document}